\documentclass[conference]{IEEEtran}
\usepackage{cite}
\usepackage{epsfig} 
\usepackage{amsmath,amssymb,amsfonts}
\usepackage{algorithmic}
\usepackage{graphicx}
\usepackage{textcomp}
\usepackage{hyperref}
\usepackage{makecell}
\usepackage{multicol}
\usepackage{xcolor}
\usepackage{multirow}
\usepackage{float}
\usepackage{subcaption}
\def\BibTeX{{\rm B\kern-.05em{\sc i\kern-.025em b}\kern-.08em
    T\kern-.1667em\lower.7ex\hbox{E}\kern-.125emX}}
\begin{document}

\title{Face Feature Visualisation of Single Morphing Attack Detection}

\thanks{This work is supported by the European Union’s Horizon 2020 research and innovation program under grant agreement No 883356 and the German Federal Ministry of Education and Research and the Hessian Ministry of Higher Education, Research, Science and the Arts within their joint support of the National Research Center for Applied Cybersecurity ATHENE.}

\author{
\IEEEauthorblockN{ Juan~E.~Tapia, Christoph Busch}
\IEEEauthorblockA{\textit{da/sec-Biometrics and Internet Security Research Group, }
\textit{Hochschule Darmstadt, Germany}\\
juan.tapia-farias, christoph.busch@h-da.de}
}

\IEEEoverridecommandlockouts
\IEEEpubid{\makebox[\columnwidth]{979-8-3503-3607-8/23/\$31.00 ©2023 IEEE \hfill} \hspace{\columnsep}\makebox[\columnwidth]{ }}

\maketitle

\IEEEpubidadjcol

\begin{abstract}
This paper proposes an explainable visualisation of different face feature extraction algorithms that enable the detection of bona fide and morphing images for single morphing attack detection. The feature extraction is based on raw image, shape, texture, frequency and compression. This visualisation may help to develop a Graphical User Interface for border policies and specifically for border guard personnel that have to investigate details of suspect images. A Random forest classifier was trained in a leave-one-out protocol on three landmarks-based face morphing methods and a StyleGAN-based morphing method for which morphed images are available in the FRLL database. For morphing attack detection, the Discrete Cosine-Transformation-based method obtained the best results for synthetic images and BSIF for landmark-based image features.
\end{abstract}


\section{Introduction}
\label{sec:introduction}
\IEEEPARstart{M}{orphing} can be understood as a technique to combine two o more look-alike facial images from one subject and an accomplice, who could apply for a valid passport exploiting the accomplice's identity. Morphing takes place in the enrolment process stage. The threat of morphing attacks is known for border crossing or identification control scenarios. It can be broadly divided into two types: (1) Single Image Morphing Attack Detection (S-MAD) techniques and Differential Morphing Attack Detection (D-MAD) methods. The  decision on S-MAD must be taken on a single image without a trusted image available for the same subject \cite{VenkateshSurvey}. 

S-MAD can be realised to perform both landmark-based morphing and synthetic GAN-based morphing methods \cite{MIPGAN} while integrating a variety of different feature extraction methods observing texture, shape, quality, residual noise, and others \cite{VenkateshSurvey, TapiaSMAD}.

In an operational scenario, a manual passport or ID card inspection complements this automatic classification at border control gates to detect morphing attacks when a suspicious image is presented or the border police have received previous advice. Understanding how a machine learning system can deal with modified images, how it can detect a morphed face image and how it can finally identify the most relevant facial areas is crucial. In summary, the main contributions of this paper are:

\begin{itemize}
    \item  A comprehensive analysis of several feature extraction methods for S-MAD is performed based on raw, texture, shape, frequency and compression.
    \item  An exhaustive evaluation was proposed based on leave-one-out protocols considering state-of-the-art databases. 
    \item A visualisation of all the features extracted is included in order to understand the differences between bona fide and morph images.
\end{itemize}

The rest of the article is organised as follows: Section~\ref{sec:relate} summarises the related works on PAD. The database description is explained in Section~\ref{sec:database}. The metrics are explained in Section~\ref{sec:metric}. The experiment and results framework is then presented in Section~\ref{sec:exp_results}. We conclude the article in Section~\ref{sec:conclusions}.

\section{RELATED WORK}
\label{sec:relate}

Developing robust morphing attack detection is an urgent need for deployed face recognition systems, to mitigate the risk posed by morphing attacks. For S-MAD several approaches can be found in the literature regarding "Implicit algorithms" based on traditional machine learning techniques. On the other hand, "Explicit algorithms" based on deep learning methods can reach impressive results in third-party evaluation platforms such as NIST \cite{Ngan2020FaceRV} and BOEP \cite{BOEP}. Today detection results based on an intra-dataset evaluation protocol can be obtained at low error rates. However, we can still observe open challenges in S-MAD that measuring real generalisation capabilities. The following issues can be determined according to the literature:  Cross-Dataset (CD), Cross-Morphing (CM), and Leave-One-Out (LOO) evaluations and generalisation are still challenging. Furthermore, the explainability and visualisation of the MAD algorithm are considered relevant support for border police in order to understand which areas are the most pertinent or which details a human observer can look for to detect a morphing image. 

The image texture is considered one of the most important characteristics since the analysis of face textures can be used to support fundamental image processing tasks. According to a survey by Venkatesh et al. \cite{VenkateshSurvey}, one of the first texture features-based approaches was presented by Raghavendra et al. \cite{RaghavendraDetecting}, who worked with Binarised Statistical Image Feature (BSIF). Other examples include Local Binary Patterns (LBP)~\cite{SpreeuwersEvaluation}. Regarding shape features, where the focus is more on analysing the changes of information, a Histogram of Oriented Gradients (HOG) was applied in~\cite{ScherhagPerformance, TapiaSMAD} for S-MAD. 

Some studies suggest applying image forensics techniques to detect the origin of image manipulation. They focus on noise patterns by analysing pixel discontinuities that may be impacted by morphing algorithms – like Photo Response Non-Uniformity (PRNU) \cite{ScherhagPRNU} and Sensor Pattern Noise (SPN) \cite{ZhangFS-SPN}, or on image quality by quantifying image degradation of artefacts in morphed faces~\cite{KraetzerForensics}. 

Tapia et al.~\cite{TapiaSMAD} proposed adding an extra Feature Selection (FS) stage after feature extraction of LBP, HOG and Raw images based on Mutual Information. Since high redundancy between features confuses the classifier, they identify the most relevant features, and remove the most redundant ones from the feature vector, to better separate bona fide and morphed images in an S-MAD scenario. The authors also conclude that the eyes and nose are the most relevant facial areas.

Very recently, Dargaud et al. ~\cite{Dargaud_2023_WACV} proposed a visualisation approach based on 50 Principal Component Analyses (PCA) and explored several colour channels, such as RGB, HSV and others, to determine that the blue channel is one of the most relevant to visualise the difference between bona fide and morphed images.

\section{Databases}
\label{sec:database}

In this study, four different databases of frontal faces images are used: the Facial Recognition Technology (FERET)~\cite{Phillips-FERET-1998}, the Face Recognition Grand Challenge (FRGCv2)~\cite{PhillipsFRGC}, the  Face Research London Lab (FRLL)\cite{FRLL} and AMSL database \cite{debruine2017face}. The morphed images in these datasets have been created using a morphing factor of 0.5, meaning both parent images contribute equally to the morphed image. FERET and FRGCv2 morphed images have been used to complement bona fide images. The FRLL and AMSL morphed images have been generated with the following morphing tools: FaceMorpher, FaceFusion and Webmorpher based on landmarks and StyleGAN from FRLL without landmarks. Table \ref{tb:databases} provides a summary of the datasets. 

FERET dataset is a subset of the Colour FERET Database, generated in the context of the Facial Recognition Technology program technically handled by the National Institute of Standards and Technology (NIST). It contains 569 faces bona fide.

The FRGCv2 dataset used in this work is a constrained subset of the second version of the Face Recognition Grand Challenge dataset. It contains 979 bona fide face images.

The FRLL dataset is a subset of the publicly available Face Research London Lab dataset. It contains 102 bona fide neutral and 102 smiling images. Three morphing algorithms were applied to obtain 1,222 morphs from the FaceMorpher algorithm, 1,222 morphs from the StyleGAN algorithm, and 1,222 morphs from the WebMorph algorithm \cite{webmorph}.

The AMSL Face Morph Image Data Set is a collection of bona fide and morphed face images that can be used to evaluate the detection performance of MAD algorithms. The images are organised as follows: genuine-neutral with 102 genuine neutral face images, genuine-smiling with 102 genuine smiling face images and 2,175 morphing face images. The following morphing tool has been used:

\begin{itemize}
    \item \textbf{FaceFusion}~\cite{facefusion}: this proprietary mobile application developed by MOMENT generates realistic faces since morphing artefacts are almost invisible.
    \item \textbf{FaceMorpher}~\cite{facemorpher}: this open-source Python implementation relies on STASM, a facial feature finding the package, for landmark detection, but generated morphs show many artefacts which make them more recognisable.
    \item \textbf{OpenCV}~\cite{opencv_library}: this open-source morphing algorithm is similar to the FaceMorpher method, but it uses Dlib to detect face landmarks. Again, some artefacts remain in generated morphs.
    \item \textbf{Webmorpher}~\cite{webmorph}: this open-source morphing algorithm is a web-based version of Psychomorph with several additional functions. While WebMorph is optimized for averaging and transforming faces, you can delineate and average any image.
    \item \textbf{StyleGAN2}~\cite{stylegan}: this open-source morphing algorithm by NVIDIA, No landmarks are used.
\end{itemize}

\begin{table}[H]
\centering
\scriptsize
\caption{Summary databases. MT: Morphing Tools.}
\label{tb:databases}
\begin{tabular}{|c|c|c|c|}
\hline
Dataset               & Bona fide           & Morphing & Notes \\ \hline
AMSL                  & 204                 & 2,175    & \begin{tabular}[c]{@{}c@{}}* Bona fide subjects are \\ the same of FRLL\end{tabular} \\ \hline
\multirow{4}{*}{FRLL} & \multirow{4}{*}{0*} & 1,222    & MT: OpenCV \\ \cline{3-4} 
                      &                     & 1,222    & MT: FaceMorpher                                                             \\ \cline{3-4} 
                      &                     & 1,222    & MT: StyleGAN2                                                              \\ \cline{3-4} 
                      &                     & 1,222    & MT: WebMorpher                                                              \\ \hline
FERET                 & 529                 & 2,116        & N/A                                                                         \\ \hline
FRGC                  & 979                 & 3,904        & N/A                                                                         \\ \hline
Total                 & 1.718               & 13.083    & 14,801                                                                       \\ \hline
\end{tabular}
\end{table}

\section{METRICS}
\label{sec:metric}

The detection performance of the investigated S-MAD algorithms was measured according to ISO/IEC 30107-3 \footnote{\url{hhttps://www.iso.org/standard/79520.html}} using the Equal Error Rate (EER), Bona fide Presentation Classification Error Rate (BPCER), and Attack Presentation Classification Error Rate (APCER) metric defined as (\ref{eq:bpcer}) and (\ref{eq:apcer}).

\begin{equation}\label{eq:bpcer}
    BPCER=\frac{\sum_{i=1}^{N_{BF}}RES_{i}}{N_{BF}}
\end{equation}

\begin{equation}\label{eq:apcer}
    APCER=\frac{1}{N_{PAIS}}\sum_{i=1}^{N_{PAIS}}(1-RES_{i})
\end{equation}

where $N_{BF}$ is the number of bona fide presentations, $N_{PAIS}$ is the number of morphing attacks for a given attack instrument species and $RES_{i}$ is $1$ if the system's response to the $i-th$ attack is classified as an attack and $0$ if classified as bona fide. In this work, S-MAD performance is reported using the Equal Error Rate (EER), which is the point where the APCER is equal to BPCER. Also, two operational points are reported BPCER10 and BPCER20. The BPCER20 is the BPCER value obtained when the APCER is fixed at 5\%, and BPCER10 (APCER at 10\%).

\section{METHOD}
\label{sec:method}

\subsection{Feature extraction methods}

Our intention is to determine which feature would be the most useful and can deliver the most specific information to separate both classes. In order to obtain and leverage different features for the bona fide and morphing images, eight different feature extraction methods and several combinations are utilised: RAW images (Intensity levels), Discrete Fourier Transform (DFT)~\cite{tan2013digital}, Steganalysis Rich Model (SRM)~\cite{zhou2018learning}, Error Level Analysis (ELA)~\cite{krawetz2007picture}, Single Value Decomposition (SVD), Local Binary Patterns (LBP), Binary Statistical Image Feature (BISF). These methods are used separately as input for the Random Forest Classifier and tested against each other.

The purpose of the DFT~\cite{tan2013digital} is to transform the image into its frequency domain representation. The intuition behind this is that differences between the frequencies of multiple face capture devices, which were used to generate the parent images.

SRM has been used successfully in related works~\cite{zhou2018learning} in order to detect MA by utilising the local noise features in an image. 

ELA is used to detect differences of compression in distinct regions of a JPEG image, which may be a residual effect caused by tampering with an image in JPEG format and resaving it in that same format. 

These feature extraction methods are applied to the original $1280\times 720$ resolution image, which is then resized and cropped to the input shape of the network.  This specific order of preprocessing contributes to a better separation of the classes, whereas resizing the image and extracting the features resulted in worse classification performance in all tests.

\subsubsection{Intensity}

For raw data, the intensity of the values in grayscale was used and normalised between 0 and 1. 

\subsubsection{Discrete Fourier transform}

The discrete Fourier transform (DFT) decomposes a discrete time-domain signal into its frequency components. 


For training purposes, only the magnitude (real) and not the phase (complex) is used. The magnitude image is then transformed from a linear scale to a logarithmic scale to compress the range of values. Furthermore, the quadrants of the matrix are shifted so that zero-value frequencies are placed at the centre of the image. 

\subsubsection{Uniform Local Binary Pattern}

The histogram of uniform LBP and BSIF ~\cite{LBPs} were used for texture. LBP is a grey-scale texture operator which characterises the spatial structure of the local image texture. Given a central pixel in the image, a binary pattern number is computed by comparing its value with those of its neighbours. 



\subsubsection {The Binary Statistical Image Feature} was also explored as a texture method. BSIF is a local descriptor designed by binarising the responses to linear filters.  The filters learn from 13 natural images. The code value of pixels is considered a local descriptor of the image intensity pattern in the pixels’ surroundings. The value of each element (i.e bit) in the binary code string is computed by binarising the response of a linear filter with a zero threshold. Each bit is associated with a different filter, and the length of the bit string determines the number of filters used. A grid search from the 60 filters available in BSIF implementation was explored. The filter $5\times5$ and 9 bits obtained the best results estimated from the baseline approach. The resulting BSIF images were used as input for the classifiers.

\subsubsection{Inverse Histogram Oriented Gradient}

For the purpose of describing the shape, the inverse Histogram of oriented gradients \cite{iHOG} was used. The distribution directions of gradients are used as features. Gradients, $x$, and $y$ derivatives of an image are helpful because the magnitude of gradients is large around edges and corners (regions of abrupt intensity changes). We know edges and corners contain more information about object shapes than flat regions. We used the visualisation proposed by Vondrik et al. \cite{iHOG} to select the best parameters that allow us to visualise the artefacts contained in morphed images. This implementation used $10\times12$ blocks and $3\times3$ filter sizes.

\subsubsection{Steganalysis Rich Model}

\begin{figure}[H]
\centering
\includegraphics[scale=0.14]{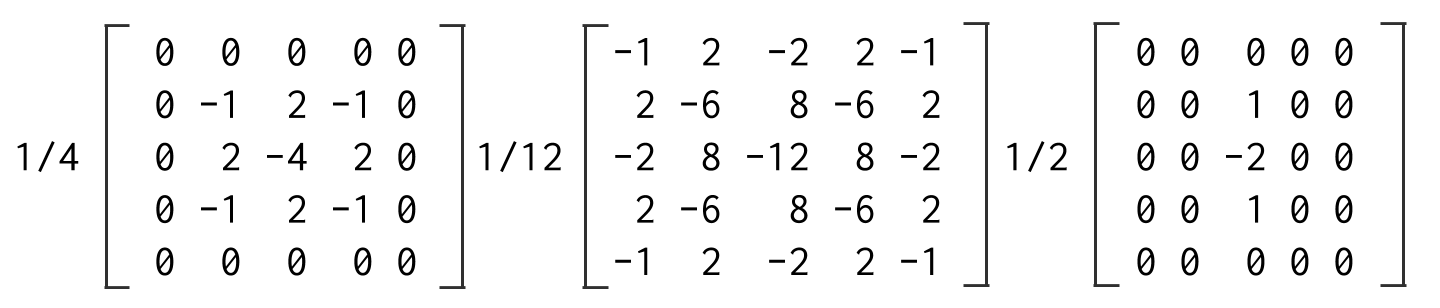}
\caption{SRM filter kernels.}
\label{fig:srmkernel}
\end{figure}

SRM filters yield noise features from neighbouring pixels, which can be applied to detect discrepancies between real and tampered images. The input and output are 3-channel images. As used by Zhou et al.~\cite{zhou2018learning}, the kernels shown in Figure~\ref{fig:srmkernel} are applied to the images, which are then directly used as the input for training the networks.

\subsubsection{Error level Analysis}
ELA~\cite{krawetz2007picture,luo2010jpeg} is a forensic method to identify portions of an image saved in a JPEG format with a different level of compression. ELA is based on characteristics of image formats that are based on a lossy image compression technique that could be used to determine if an image has been digitally modified. 

JPEG is a method of lossy compression for digital images. The compression level is chosen as a trade-off between image size and quality. A JPEG compression scale is usually 70\%. The compression of data discards or loses information.
The JPEG algorithm works on image grids, compressed independently, having a size of $8\times8$ pixels. The $8\times8$ dimension was chosen using a grid search. Meanwhile, any matrices of size less than $8\times8$ do not have enough information. They result in poor-quality compressed images.

ELA highlights differences in the JPEG compression rate. Regions with uniform colourings, like solid blue or white pixels, will likely have a lower ELA result (darker colour) than high-contrast edges. Highlighted regions can be, potentially, tampered regions in the image that suffered a second JPEG compression after the user saves the tampered image.

\begin{figure*}
\begin{centering}
\includegraphics[scale=0.4]{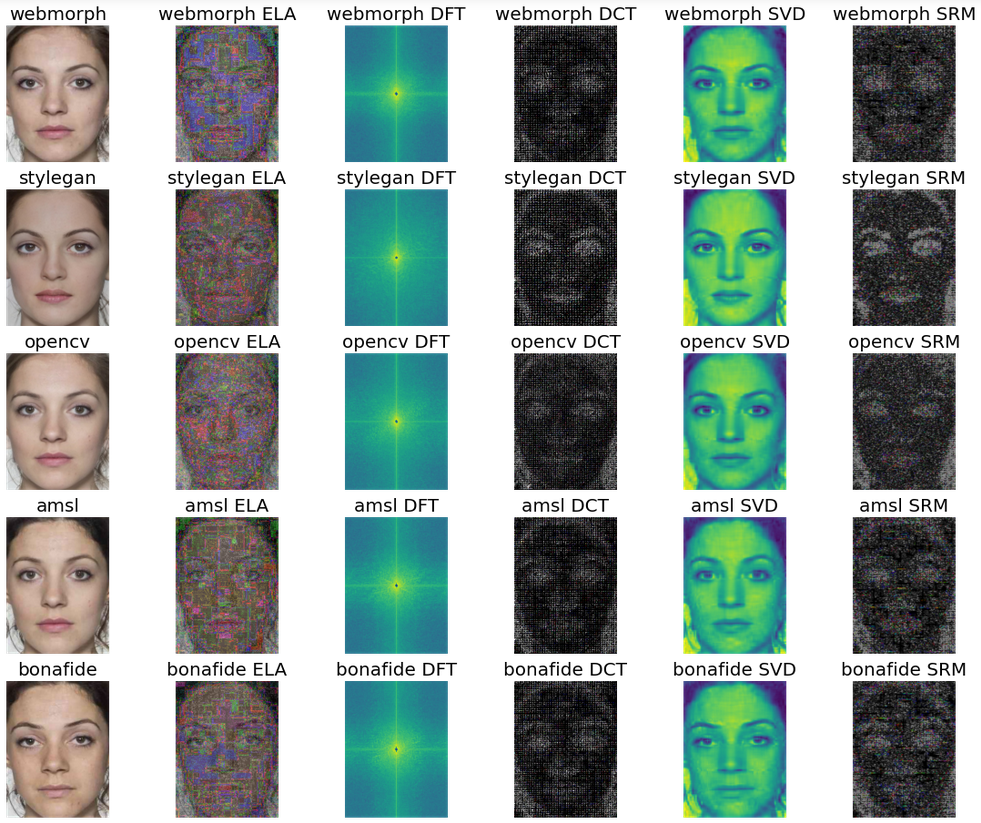}
\caption{\label{fig:examples} Examples of face feature explainability for a random bona fide and Morph.}
\par\end{centering}
\end{figure*}

Figure~\ref{fig:examples} presents visualisation examples for ELA, DFT, DCT, SVD and SRM morphed images. The last row, show visualisation examples of a random bona fide image. 

\section{EXPERIMENTS AND RESULTS}
\label{sec:exp_results}

In this paper, a LOO protocol was defined in order to evaluate the influence of the feature extracted. According to each experiment, we explored five cross-dataset tests.  In the end, we performed 125 evaluations in total. In order to train S-MAD, all the datasets were divided into 70,0\% for train and 30,0\% for testing.

A LOO protocol was applied to train and test the S-MAD system, which means in the first round, FaceFusion was used to compute the morphing images used for the test, and training was carried out with FaceMorpher, OpenCV-Morpher, UBO-Morpher and FRLL. In the second round, FaceMorpher was used for testing and training was done with morphed images created with FaceFusion, OpenCV-Morpher, UBO-Morpher and FRLL, and so on. All the images were aligned, cropped and resized to $180\times240$. 

Different kinds of features were extracted from faces based on SRM, ELA, DFT, SVD, LBP and BSIF filters for all experiments. We used the intensity values of the pixels from raw images normalised between 0 and 1. The histogram of the uniform LBP and BSIF was used for texture. For the uLBP, all radii values were explored from uLBP81 to uLBP88. The fusion of LBPs was also investigated, concatenating the LBP81 to LBP88. The image's vertical (uLBP\_VERT) and horizontal (uLBP\_HOR) concatenation divided into eight patches was also explored. 

After feature extraction, we fused that information at the feature level by concatenating the feature vectors from different sources into a single feature vector that becomes the input to the classifier.
From BSIF, the resulting images of filter $3\times 3-5bits$ were used considering BISF images (BSIF-IM), BSIF Histogram (BSIF-H) and BSIF normalised histogram (BSIF-NH).  All the features were extracted after applying our proposed transfer texture method. 

The S-MAD system was trained with a Random Forest classifier with each feature extraction method described above as input. All the individual and average results through all features extracted are presented in Table \ref{tab:sumary-table1}. The first column of Table \ref{tab:sumary-table1} identifies each evaluation's LOO dataset test.

\begin{figure*}
\begin{centering}
\includegraphics[scale=0.57]{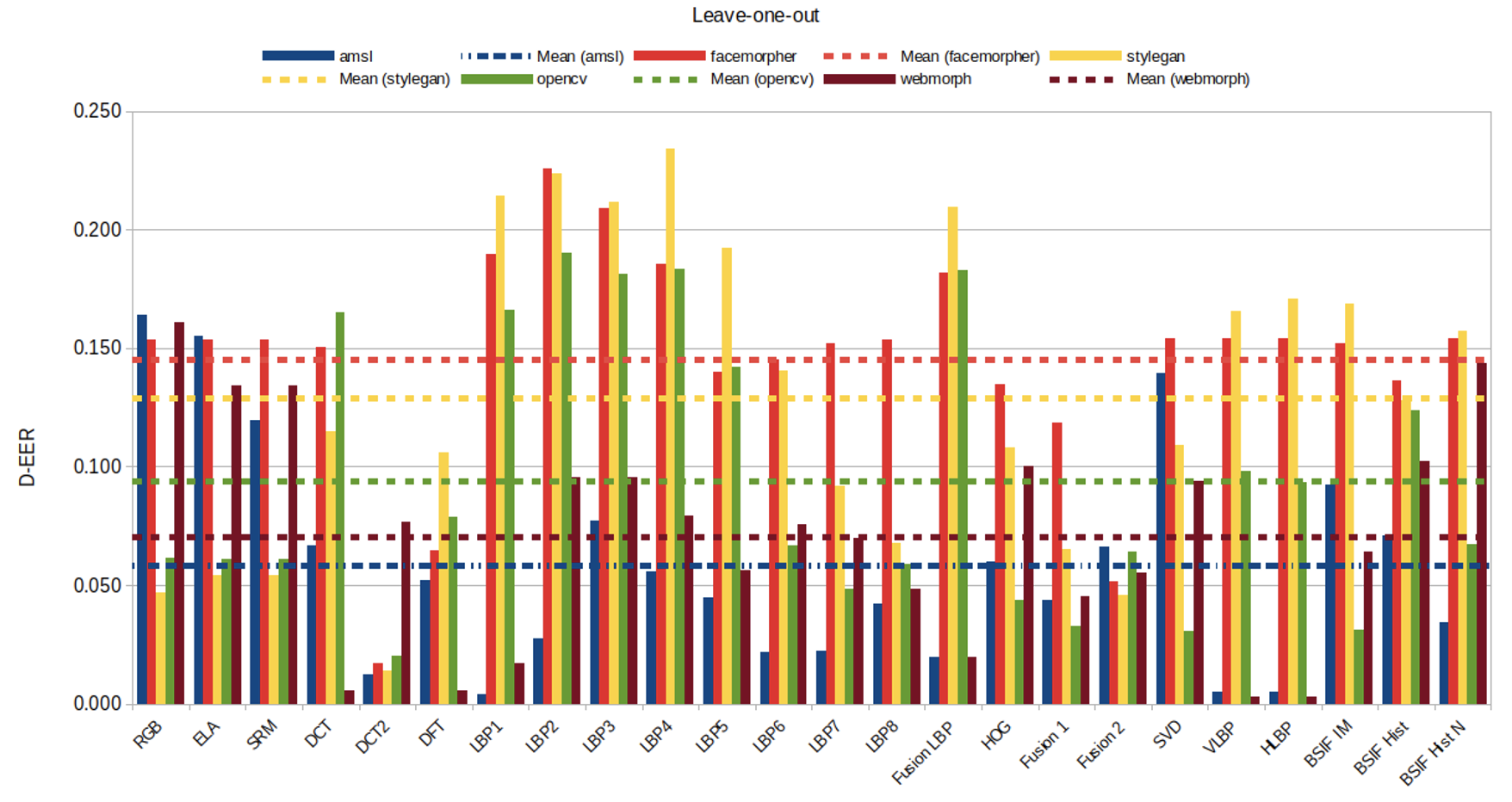}
\caption{\label{fig:plot1} Summary results of LOO protocol across all the features. The Dot line represents the average results for each dataset.}
\par\end{centering}
\end{figure*}

Figure \ref{fig:plot1} show a bar plot with all the feature methods used on X-axis and the EER on the Y-axis. A dot-plot line was added in order to show the average results of all the methods for each dataset. For the AMSL dataset reaches the lower EER. The higher EER is obtained for MA developed with FaceMorpher images. The results obtained in LBP show that this texture feature is not suitable for identifying synthetic images such as StyleGAN2 images. Conversely, DCT shows impressive results across the datasets, including the StyleGANs MA.


\begin{table*}[]
\centering
\caption{Summary results for all the features extracted on  the LOO protocol. All the results represent the EER.}
\label{tab:sumary-table1}
\resizebox{\textwidth}{!}{%
\begin{tabular}{|c|ccccccccccccccc|}
\hline
\textbf{\begin{tabular}[c]{@{}c@{}}ONE OUT: \\ Morph AMSL\end{tabular}}     & \multicolumn{15}{c|}{Algorithms (EER)}                                                                                                                                                                                                                                                                                                                                                                                                                                                                                                                                                                                                                                                                                     \\ \hline
Morphing-Test & \multicolumn{1}{c|}{RGB}            & \multicolumn{1}{c|}{ELA}   & \multicolumn{1}{c|}{SRM}   & \multicolumn{1}{c|}{DCT2}           & \multicolumn{1}{c|}{DFT}            & \multicolumn{1}{c|}{\begin{tabular}[c]{@{}c@{}}LBP\\ 8,1\end{tabular}} & \multicolumn{1}{c|}{\begin{tabular}[c]{@{}c@{}}Fusion\\  LBP\end{tabular}} & \multicolumn{1}{c|}{HOG}   & \multicolumn{1}{c|}{SVD}   & \multicolumn{1}{c|}{VLBP}           & \multicolumn{1}{c|}{HLBP}           & \multicolumn{1}{c|}{\begin{tabular}[c]{@{}c@{}}BSIF\\  IM\end{tabular}} & \multicolumn{1}{c|}{\begin{tabular}[c]{@{}c@{}}BSIF\\  Hist\end{tabular}} & \multicolumn{1}{c|}{\begin{tabular}[c]{@{}c@{}}BSIF\\  Hist N\end{tabular}} & Average        \\ \hline
FaceMorpher                                                                 & \multicolumn{1}{c|}{0.377}          & \multicolumn{1}{c|}{0.357} & \multicolumn{1}{c|}{0.498} & \multicolumn{1}{c|}{0.007}          & \multicolumn{1}{c|}{0.035}          & \multicolumn{1}{c|}{0.002}                                             & \multicolumn{1}{c|}{0.052}                                                 & \multicolumn{1}{c|}{0.453} & \multicolumn{1}{c|}{0.335} & \multicolumn{1}{c|}{0.002}          & \multicolumn{1}{c|}{0.003}          & \multicolumn{1}{c|}{0.345}                                              & \multicolumn{1}{c|}{0.278}                                                & \multicolumn{1}{c|}{0.119}                                                  & 0.196          \\ \hline
OpenCV                                                                      & \multicolumn{1}{c|}{0.332}          & \multicolumn{1}{c|}{0.325} & \multicolumn{1}{c|}{0.489} & \multicolumn{1}{c|}{0.014}          & \multicolumn{1}{c|}{0.104}          & \multicolumn{1}{c|}{0.009}                                             & \multicolumn{1}{c|}{0.014}                                                 & \multicolumn{1}{c|}{0.349} & \multicolumn{1}{c|}{0.321} & \multicolumn{1}{c|}{0.003}          & \multicolumn{1}{c|}{0.003}          & \multicolumn{1}{c|}{0.333}                                              & \multicolumn{1}{c|}{0.184}                                                & \multicolumn{1}{c|}{0.084}                                                  & 0.186          \\ \hline
StyleGAN2                                                                    & \multicolumn{1}{c|}{0.030}          & \multicolumn{1}{c|}{0.035} & \multicolumn{1}{c|}{0.465} & \multicolumn{1}{c|}{0.041}          & \multicolumn{1}{c|}{0.151}          & \multicolumn{1}{c|}{0.012}                                             & \multicolumn{1}{c|}{0.048}                                                 & \multicolumn{1}{c|}{0.319} & \multicolumn{1}{c|}{0.066} & \multicolumn{1}{c|}{0.008}          & \multicolumn{1}{c|}{0.008}          & \multicolumn{1}{c|}{0.239}                                              & \multicolumn{1}{c|}{0.114}                                                & \multicolumn{1}{c|}{0.033}                                                  & 0.151          \\ \hline
Webmorph                                                                    & \multicolumn{1}{c|}{0.023}          & \multicolumn{1}{c|}{0.026} & \multicolumn{1}{c|}{0.165} & \multicolumn{1}{c|}{0.003}          & \multicolumn{1}{c|}{0.012}          & \multicolumn{1}{c|}{0.003}                                             & \multicolumn{1}{c|}{0.007}                                                 & \multicolumn{1}{c|}{0.244} & \multicolumn{1}{c|}{0.032} & \multicolumn{1}{c|}{0.017}          & \multicolumn{1}{c|}{0.017}          & \multicolumn{1}{c|}{0.070}                                              & \multicolumn{1}{c|}{0.064}                                                & \multicolumn{1}{c|}{0.032}                                                  & \textbf{0.065} \\ \hline
AMSL-Avg.                                                                   & \multicolumn{1}{c|}{0.164}          & \multicolumn{1}{c|}{0.155} & \multicolumn{1}{c|}{0.120} & \multicolumn{1}{c|}{\textbf{0.012}} & \multicolumn{1}{c|}{0.052}          & \multicolumn{1}{c|}{\textbf{0.004}}                                    & \multicolumn{1}{c|}{0.020}                                                 & \multicolumn{1}{c|}{0.060} & \multicolumn{1}{c|}{0.139} & \multicolumn{1}{c|}{\textbf{0.005}} & \multicolumn{1}{c|}{\textbf{0.005}} & \multicolumn{1}{c|}{0.092}                                              & \multicolumn{1}{c|}{0.071}                                                & \multicolumn{1}{c|}{0.034}                                                  & \textbf{0.058} \\ \hline
\textbf{\begin{tabular}[c]{@{}c@{}}ONE OUT: \\ FaceMorpher\end{tabular}}    & \multicolumn{15}{c|}{Algorithms (EER)}                                                                                                                                                                                                                                                                                                                                                                                                                                                                                                                                                                                                                                                                                     \\ \hline
Morphing-Test                                                               & \multicolumn{1}{c|}{RGB}            & \multicolumn{1}{c|}{ELA}   & \multicolumn{1}{c|}{SRM}   & \multicolumn{1}{c|}{DCT2}           & \multicolumn{1}{c|}{DFT}            & \multicolumn{1}{c|}{\begin{tabular}[c]{@{}c@{}}LBP\\ 8,1\end{tabular}} & \multicolumn{1}{c|}{\begin{tabular}[c]{@{}c@{}}Fusion \\ LBP\end{tabular}} & \multicolumn{1}{c|}{HOG}   & \multicolumn{1}{c|}{SVD}   & \multicolumn{1}{c|}{VLBP}           & \multicolumn{1}{c|}{HLBP}           & \multicolumn{1}{c|}{\begin{tabular}[c]{@{}c@{}}BSIF\\  IM\end{tabular}} & \multicolumn{1}{c|}{\begin{tabular}[c]{@{}c@{}}BSIF\\  Hist\end{tabular}} & \multicolumn{1}{c|}{\begin{tabular}[c]{@{}c@{}}BSIF \\ Hist N\end{tabular}} & Average        \\ \hline
AMSL                                                                        & \multicolumn{1}{c|}{0.50}           & \multicolumn{1}{c|}{0.50}  & \multicolumn{1}{c|}{0.50}  & \multicolumn{1}{c|}{0.09}           & \multicolumn{1}{c|}{0.30}           & \multicolumn{1}{c|}{0.39}                                              & \multicolumn{1}{c|}{0.42}                                                  & \multicolumn{1}{c|}{0.50}  & \multicolumn{1}{c|}{0.50}  & \multicolumn{1}{c|}{0.50}           & \multicolumn{1}{c|}{0.50}           & \multicolumn{1}{c|}{0.50}                                               & \multicolumn{1}{c|}{0.41}                                                 & \multicolumn{1}{c|}{0.50}                                                   & 0.449          \\ \hline
OpenCV                                                                      & \multicolumn{1}{c|}{0.09}           & \multicolumn{1}{c|}{0.09}  & \multicolumn{1}{c|}{0.10}  & \multicolumn{1}{c|}{0.04}           & \multicolumn{1}{c|}{0.14}           & \multicolumn{1}{c|}{0.04}                                              & \multicolumn{1}{c|}{0.03}                                                  & \multicolumn{1}{c|}{0.14}  & \multicolumn{1}{c|}{0.09}  & \multicolumn{1}{c|}{0.09}           & \multicolumn{1}{c|}{0.09}           & \multicolumn{1}{c|}{0.09}                                               & \multicolumn{1}{c|}{0.07}                                                 & \multicolumn{1}{c|}{0.09}                                                   & \textbf{0.091} \\ \hline
StyleGAN2                                                                    & \multicolumn{1}{c|}{0.50}           & \multicolumn{1}{c|}{0.50}  & \multicolumn{1}{c|}{0.10}  & \multicolumn{1}{c|}{0.09}           & \multicolumn{1}{c|}{0.31}           & \multicolumn{1}{c|}{0.03}                                              & \multicolumn{1}{c|}{0.13}                                                  & \multicolumn{1}{c|}{0.49}  & \multicolumn{1}{c|}{0.50}  & \multicolumn{1}{c|}{0.50}           & \multicolumn{1}{c|}{0.50}           & \multicolumn{1}{c|}{0.50}                                               & \multicolumn{1}{c|}{0.24}                                                 & \multicolumn{1}{c|}{0.50}                                                   & 0.344          \\ \hline
WebMorph                                                                    & \multicolumn{1}{c|}{0.50}           & \multicolumn{1}{c|}{0.50}  & \multicolumn{1}{c|}{0.50}  & \multicolumn{1}{c|}{0.09}           & \multicolumn{1}{c|}{0.33}           & \multicolumn{1}{c|}{0.44}                                              & \multicolumn{1}{c|}{0.47}                                                  & \multicolumn{1}{c|}{0.50}  & \multicolumn{1}{c|}{0.50}  & \multicolumn{1}{c|}{0.50}           & \multicolumn{1}{c|}{0.50}           & \multicolumn{1}{c|}{0.49}                                               & \multicolumn{1}{c|}{0.45}                                                 & \multicolumn{1}{c|}{0.50}                                                   & 0.459          \\ \hline
FaceMorpher-Avg.                                                            & \multicolumn{1}{c|}{0.153}          & \multicolumn{1}{c|}{0.154} & \multicolumn{1}{c|}{0.154} & \multicolumn{1}{c|}{\textbf{0.017}} & \multicolumn{1}{c|}{0.065}          & \multicolumn{1}{c|}{0.190}                                             & \multicolumn{1}{c|}{0.182}                                                 & \multicolumn{1}{c|}{0.135} & \multicolumn{1}{c|}{0.154} & \multicolumn{1}{c|}{0.154}          & \multicolumn{1}{c|}{0.154}          & \multicolumn{1}{c|}{0.152}                                              & \multicolumn{1}{c|}{0.136}                                                & \multicolumn{1}{c|}{0.154}                                                  & 0.145          \\ \hline
\textbf{\begin{tabular}[c]{@{}c@{}}ONE OUT: \\ OpenCV\end{tabular}}         & \multicolumn{15}{c|}{Algorithms (EER)}                                                                                                                                                                                                                                                                                                                                                                                                                                                                                                                                                                                                                                                                                     \\ \hline
Morphing-Test                                                               & \multicolumn{1}{c|}{RGB}            & \multicolumn{1}{c|}{ELA}   & \multicolumn{1}{c|}{SRM}   & \multicolumn{1}{c|}{DCT2}           & \multicolumn{1}{c|}{DFT}            & \multicolumn{1}{c|}{\begin{tabular}[c]{@{}c@{}}LBP\\ 8,1\end{tabular}} & \multicolumn{1}{c|}{\begin{tabular}[c]{@{}c@{}}Fusion\\  LBP\end{tabular}} & \multicolumn{1}{c|}{HOG}   & \multicolumn{1}{c|}{SVD}   & \multicolumn{1}{c|}{VLBP}           & \multicolumn{1}{c|}{HLBP}           & \multicolumn{1}{c|}{\begin{tabular}[c]{@{}c@{}}BSIF\\  IM\end{tabular}} & \multicolumn{1}{c|}{\begin{tabular}[c]{@{}c@{}}BSIF\\  Hist\end{tabular}} & \multicolumn{1}{c|}{\begin{tabular}[c]{@{}c@{}}BSIF\\  Hist N\end{tabular}} & Average        \\ \hline
AMSL                                                                        & \multicolumn{1}{c|}{0.124}          & \multicolumn{1}{c|}{0.146} & \multicolumn{1}{c|}{0.477} & \multicolumn{1}{c|}{0.048}          & \multicolumn{1}{c|}{0.191}          & \multicolumn{1}{c|}{0.500}                                             & \multicolumn{1}{c|}{0.500}                                                 & \multicolumn{1}{c|}{0.495} & \multicolumn{1}{c|}{0.222} & \multicolumn{1}{c|}{0.357}          & \multicolumn{1}{c|}{0.371}          & \multicolumn{1}{c|}{0.440}                                              & \multicolumn{1}{c|}{0.221}                                                & \multicolumn{1}{c|}{0.322}                                                  & 0.339          \\ \hline
FaceMorpher                                                                 & \multicolumn{1}{c|}{0.000}          & \multicolumn{1}{c|}{0.000} & \multicolumn{1}{c|}{0.121} & \multicolumn{1}{c|}{0.011}          & \multicolumn{1}{c|}{0.027}          & \multicolumn{1}{c|}{0.143}                                             & \multicolumn{1}{c|}{0.129}                                                 & \multicolumn{1}{c|}{0.194} & \multicolumn{1}{c|}{0.004} & \multicolumn{1}{c|}{0.001}          & \multicolumn{1}{c|}{0.000}          & \multicolumn{1}{c|}{0.000}                                              & \multicolumn{1}{c|}{0.001}                                                & \multicolumn{1}{c|}{0.000}                                                  & \textbf{0.064} \\ \hline
StyleGAN2                                                                    & \multicolumn{1}{c|}{0.149}          & \multicolumn{1}{c|}{0.172} & \multicolumn{1}{c|}{0.073} & \multicolumn{1}{c|}{0.028}          & \multicolumn{1}{c|}{0.003}          & \multicolumn{1}{c|}{0.001}                                             & \multicolumn{1}{c|}{0.034}                                                 & \multicolumn{1}{c|}{0.456} & \multicolumn{1}{c|}{0.410} & \multicolumn{1}{c|}{0.096}          & \multicolumn{1}{c|}{0.095}          & \multicolumn{1}{c|}{0.497}                                              & \multicolumn{1}{c|}{0.059}                                                & \multicolumn{1}{c|}{0.102}                                                  & 0.143          \\ \hline
Webmorph                                                                    & \multicolumn{1}{c|}{0.103}          & \multicolumn{1}{c|}{0.116} & \multicolumn{1}{c|}{0.490} & \multicolumn{1}{c|}{0.047}          & \multicolumn{1}{c|}{0.263}          & \multicolumn{1}{c|}{0.500}                                             & \multicolumn{1}{c|}{0.500}                                                 & \multicolumn{1}{c|}{0.495} & \multicolumn{1}{c|}{0.191} & \multicolumn{1}{c|}{0.402}          & \multicolumn{1}{c|}{0.406}          & \multicolumn{1}{c|}{0.411}                                              & \multicolumn{1}{c|}{0.350}                                                & \multicolumn{1}{c|}{0.409}                                                  & 0.341          \\ \hline
OpenCV-Avg.                                                                 & \multicolumn{1}{c|}{\textbf{0.047}} & \multicolumn{1}{c|}{0.054} & \multicolumn{1}{c|}{0.054} & \multicolumn{1}{c|}{\textbf{0.014}} & \multicolumn{1}{c|}{0.106}          & \multicolumn{1}{c|}{0.214}                                             & \multicolumn{1}{c|}{0.209}                                                 & \multicolumn{1}{c|}{0.108} & \multicolumn{1}{c|}{0.109} & \multicolumn{1}{c|}{0.165}          & \multicolumn{1}{c|}{0.171}          & \multicolumn{1}{c|}{0.168}                                              & \multicolumn{1}{c|}{0.128}                                                & \multicolumn{1}{c|}{0.157}                                                  & 0.128          \\ \hline
\textbf{\begin{tabular}[c]{@{}c@{}}ONE OUT: \\ Morph StyleGAN2\end{tabular}} & \multicolumn{15}{c|}{Algorithms (EER)}                                                                                                                                                                                                                                                                                                                                                                                                                                                                                                                                                                                                                                                                                     \\ \hline
Morphing-Test                                                               & \multicolumn{1}{c|}{RGB}            & \multicolumn{1}{c|}{ELA}   & \multicolumn{1}{c|}{SRM}   & \multicolumn{1}{c|}{DCT2}           & \multicolumn{1}{c|}{DFT}            & \multicolumn{1}{c|}{\begin{tabular}[c]{@{}c@{}}LBP\\ 8,1\end{tabular}} & \multicolumn{1}{c|}{\begin{tabular}[c]{@{}c@{}}Fusion\\  LBP\end{tabular}} & \multicolumn{1}{c|}{HOG}   & \multicolumn{1}{c|}{SVD}   & \multicolumn{1}{c|}{VLBP}           & \multicolumn{1}{c|}{HLBP}           & \multicolumn{1}{c|}{\begin{tabular}[c]{@{}c@{}}BSIF\\  IM\end{tabular}} & \multicolumn{1}{c|}{\begin{tabular}[c]{@{}c@{}}BSIF\\  Hist\end{tabular}} & \multicolumn{1}{c|}{\begin{tabular}[c]{@{}c@{}}BSIF\\  Hist N\end{tabular}} & Average        \\ \hline
AMSL                                                                        & \multicolumn{1}{c|}{0.410}          & \multicolumn{1}{c|}{0.408} & \multicolumn{1}{c|}{0.281} & \multicolumn{1}{c|}{0.036}          & \multicolumn{1}{c|}{0.170}          & \multicolumn{1}{c|}{0.500}                                             & \multicolumn{1}{c|}{0.500}                                                 & \multicolumn{1}{c|}{0.495} & \multicolumn{1}{c|}{0.342} & \multicolumn{1}{c|}{0.168}          & \multicolumn{1}{c|}{0.171}          & \multicolumn{1}{c|}{0.252}                                              & \multicolumn{1}{c|}{0.218}                                                & \multicolumn{1}{c|}{0.223}                                                  & 0.335          \\ \hline
FaceMorpher                                                                 & \multicolumn{1}{c|}{0.299}          & \multicolumn{1}{c|}{0.295} & \multicolumn{1}{c|}{0.153} & \multicolumn{1}{c|}{0.077}          & \multicolumn{1}{c|}{0.315}          & \multicolumn{1}{c|}{0.500}                                             & \multicolumn{1}{c|}{0.255}                                                 & \multicolumn{1}{c|}{0.461} & \multicolumn{1}{c|}{0.276} & \multicolumn{1}{c|}{0.018}          & \multicolumn{1}{c|}{0.029}          & \multicolumn{1}{c|}{0.319}                                              & \multicolumn{1}{c|}{0.054}                                                & \multicolumn{1}{c|}{0.189}                                                  & 0.226          \\ \hline
OpenCV                                                                      & \multicolumn{1}{c|}{0.259}          & \multicolumn{1}{c|}{0.265} & \multicolumn{1}{c|}{0.106} & \multicolumn{1}{c|}{0.004}          & \multicolumn{1}{c|}{0.029}          & \multicolumn{1}{c|}{0.058}                                             & \multicolumn{1}{c|}{0.014}                                                 & \multicolumn{1}{c|}{0.365} & \multicolumn{1}{c|}{0.274} & \multicolumn{1}{c|}{0.003}          & \multicolumn{1}{c|}{0.008}          & \multicolumn{1}{c|}{0.312}                                              & \multicolumn{1}{c|}{0.007}                                                & \multicolumn{1}{c|}{0.070}                                                  & 0.118          \\ \hline
Webmorph                                                                    & \multicolumn{1}{c|}{0.395}          & \multicolumn{1}{c|}{0.396} & \multicolumn{1}{c|}{0.201} & \multicolumn{1}{c|}{0.030}          & \multicolumn{1}{c|}{0.141}          & \multicolumn{1}{c|}{0.500}                                             & \multicolumn{1}{c|}{0.500}                                                 & \multicolumn{1}{c|}{0.488} & \multicolumn{1}{c|}{0.331} & \multicolumn{1}{c|}{0.246}          & \multicolumn{1}{c|}{0.240}          & \multicolumn{1}{c|}{0.254}                                              & \multicolumn{1}{c|}{0.338}                                                & \multicolumn{1}{c|}{0.303}                                                  & 0.328          \\ \hline
StyleGAN2-Avg.                                                               & \multicolumn{1}{c|}{0.061}          & \multicolumn{1}{c|}{0.061} & \multicolumn{1}{c|}{0.061} & \multicolumn{1}{c|}{\textbf{0.020}} & \multicolumn{1}{c|}{0.079}          & \multicolumn{1}{c|}{0.166}                                             & \multicolumn{1}{c|}{0.183}                                                 & \multicolumn{1}{c|}{0.044} & \multicolumn{1}{c|}{0.031} & \multicolumn{1}{c|}{0.098}          & \multicolumn{1}{c|}{0.093}          & \multicolumn{1}{c|}{0.031}                                              & \multicolumn{1}{c|}{0.124}                                                & \multicolumn{1}{c|}{\textbf{0.067}}                                         & \textbf{0.093} \\ \hline
\textbf{\begin{tabular}[c]{@{}c@{}}ONE OUT: \\ Morph Webmorph\end{tabular}} & \multicolumn{15}{c|}{Algorithms (EER)}                                                                                                                                                                                                                                                                                                                                                                                                                                                                                                                                                                                                                                                                                     \\ \hline
Morphing-Test                                                               & \multicolumn{1}{c|}{RGB}            & \multicolumn{1}{c|}{ELA}   & \multicolumn{1}{c|}{SRM}   & \multicolumn{1}{c|}{DCT2}           & \multicolumn{1}{c|}{DFT}            & \multicolumn{1}{c|}{\begin{tabular}[c]{@{}c@{}}LBP\\ 8,1\end{tabular}} & \multicolumn{1}{c|}{\begin{tabular}[c]{@{}c@{}}Fusion\\  LBP\end{tabular}} & \multicolumn{1}{c|}{HOG}   & \multicolumn{1}{c|}{SVD}   & \multicolumn{1}{c|}{VLBP}           & \multicolumn{1}{c|}{HLBP}           & \multicolumn{1}{c|}{\begin{tabular}[c]{@{}c@{}}BSIF\\  IM\end{tabular}} & \multicolumn{1}{c|}{\begin{tabular}[c]{@{}c@{}}BSIF\\  Hist\end{tabular}} & \multicolumn{1}{c|}{\begin{tabular}[c]{@{}c@{}}BSIF\\  Hist N\end{tabular}} & Average        \\ \hline
AMSL                                                                        & \multicolumn{1}{c|}{0.067}          & \multicolumn{1}{c|}{0.077} & \multicolumn{1}{c|}{0.126} & \multicolumn{1}{c|}{0.008}          & \multicolumn{1}{c|}{0.036}          & \multicolumn{1}{c|}{0.000}                                             & \multicolumn{1}{c|}{0.000}                                                 & \multicolumn{1}{c|}{0.173} & \multicolumn{1}{c|}{0.111} & \multicolumn{1}{c|}{0.000}          & \multicolumn{1}{c|}{0.000}          & \multicolumn{1}{c|}{0.055}                                              & \multicolumn{1}{c|}{0.043}                                                & \multicolumn{1}{c|}{0.016}                                                  & \textbf{0.061} \\ \hline
FaceMorpher                                                                 & \multicolumn{1}{c|}{0.399}          & \multicolumn{1}{c|}{0.414} & \multicolumn{1}{c|}{0.497} & \multicolumn{1}{c|}{0.012}          & \multicolumn{1}{c|}{0.050}          & \multicolumn{1}{c|}{0.004}                                             & \multicolumn{1}{c|}{0.041}                                                 & \multicolumn{1}{c|}{0.471} & \multicolumn{1}{c|}{0.337} & \multicolumn{1}{c|}{0.008}          & \multicolumn{1}{c|}{0.006}          & \multicolumn{1}{c|}{0.220}                                              & \multicolumn{1}{c|}{0.364}                                                & \multicolumn{1}{c|}{0.403}                                                  & 0.219          \\ \hline
OpenCV                                                                      & \multicolumn{1}{c|}{0.369}          & \multicolumn{1}{c|}{0.378} & \multicolumn{1}{c|}{0.467} & \multicolumn{1}{c|}{0.013}          & \multicolumn{1}{c|}{0.165}          & \multicolumn{1}{c|}{0.010}                                             & \multicolumn{1}{c|}{0.037}                                                 & \multicolumn{1}{c|}{0.357} & \multicolumn{1}{c|}{0.315} & \multicolumn{1}{c|}{0.005}          & \multicolumn{1}{c|}{0.008}          & \multicolumn{1}{c|}{0.220}                                              & \multicolumn{1}{c|}{0.326}                                                & \multicolumn{1}{c|}{0.227}                                                  & 0.214          \\ \hline
StyleGAN2                                                                    & \multicolumn{1}{c|}{0.061}          & \multicolumn{1}{c|}{0.072} & \multicolumn{1}{c|}{0.302} & \multicolumn{1}{c|}{0.026}          & \multicolumn{1}{c|}{0.227}          & \multicolumn{1}{c|}{0.016}                                             & \multicolumn{1}{c|}{0.076}                                                 & \multicolumn{1}{c|}{0.255} & \multicolumn{1}{c|}{0.166} & \multicolumn{1}{c|}{0.001}          & \multicolumn{1}{c|}{0.001}          & \multicolumn{1}{c|}{0.238}                                              & \multicolumn{1}{c|}{0.257}                                                & \multicolumn{1}{c|}{0.040}                                                  & 0.150          \\ \hline
Webmorph-Avg.                                                               & \multicolumn{1}{c|}{0.161}          & \multicolumn{1}{c|}{0.134} & \multicolumn{1}{c|}{0.134} & \multicolumn{1}{c|}{\textbf{0.076}} & \multicolumn{1}{c|}{\textbf{0.006}} & \multicolumn{1}{c|}{0.017}                                             & \multicolumn{1}{c|}{0.020}                                                 & \multicolumn{1}{c|}{0.100} & \multicolumn{1}{c|}{0.094} & \multicolumn{1}{c|}{\textbf{0.003}} & \multicolumn{1}{c|}{\textbf{0.003}} & \multicolumn{1}{c|}{0.064}                                              & \multicolumn{1}{c|}{0.102}                                                & \multicolumn{1}{c|}{0.143}                                                  & \textbf{0.070} \\ \hline
\end{tabular}%
}
\end{table*}

\begin{figure}[H]
\centering
\includegraphics[scale=0.25]{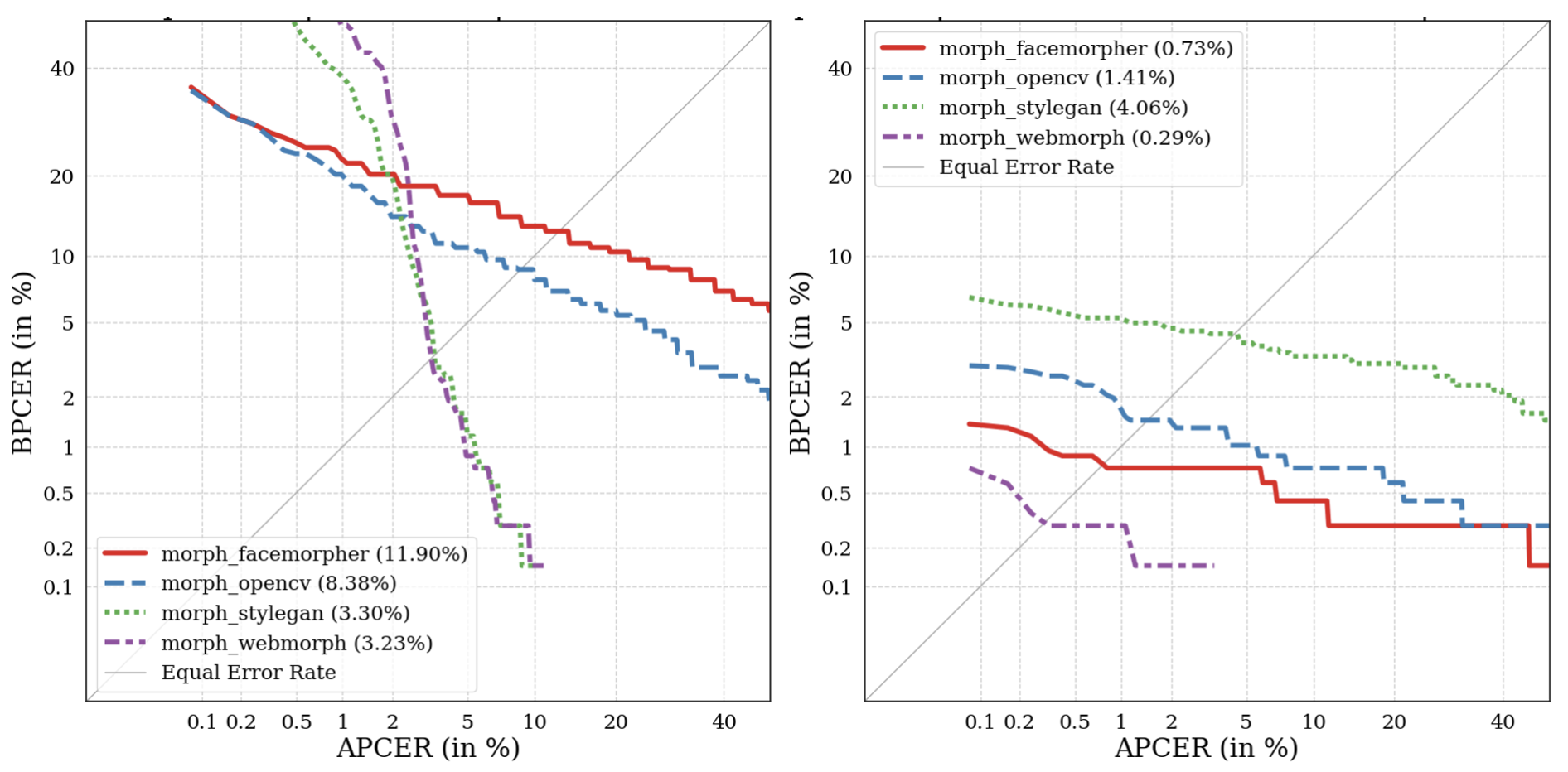}
\caption{Left. BSIF-NH - DET curve. Right: DCT DET curve. The EER is shown in parentheses in percentages.}
\label{fig:eer-plot}
\end{figure}

Figure \ref{fig:eer-plot} shows two DET plots illustrating the error trade-off for the four S-MAD methods with the EER on percentages in parentheses for the best case. The left images show the results of the DET curve for the BSIF feature on the AMSL database. Where FaceMorpher reached an EER 11.90\%, OpenCV 8.38\%, StyleGAN2 3,30\%, and WebMorph obtained a 3,23\%. The BPCER10/20 obtained is 13.5\% and 16.3\%. The right DET shows the DCT feature's results in the normalised histogram. Where FaceMorpher reached an EER 0.73\%, OpenCV 1.41\%, StyleGAN2 4.06\%, and WebMorph obtained an 0.29\%. The BPCER10/20 obtained is 0.46\% and 0.83\%.

\section{Conclusions}
\label{sec:conclusions}

This work shows that different feature extractors can deliver relevant information to guide the analysis of MAD. FaceMorpher has been identifying, on average, the morphing tool with the highest EER. Textures and frequencies are more effective in visualising the details of bona fide and morphed images without compression. ELA has been identified as a very good tool for detecting changes in JPEG compression. After this work, the main challenge is identifying common parameters to tune all the filters. As we can declare now, different datasets need proper parameters. The synthetics images based on GANs are not difficult to identify using the DCT feature compared to  Landmark-based such as FaceMorpher.  As a future work, fusion-specific features may be extended to Deep learning methods to identify specific morphing tools.

\bibliographystyle{myieee}
\bibliography{biblio}

\end{document}